\documentclass{article}

\PassOptionsToPackage{numbers, sort&compress}{natbib}

\usepackage[main, final]{neurips/neurips_2026}

\usepackage[utf8]{inputenc} 
\usepackage[T1]{fontenc}    
\usepackage{hyperref}       
\usepackage{url}            
\usepackage{booktabs}       
\usepackage{amsfonts}       
\usepackage{nicefrac}       
\usepackage{microtype}      
\usepackage{xcolor}         

\iftrue
    \newcommand{\holger}[1]{\noindent}
    \newcommand{\wang}[1]{\noindent}
    \newcommand{\lin}[1]{\noindent}
    \newcommand{\julian}[1]{\noindent}
    \newcommand{\nan}[1]{\noindent}
    \DeclareRobustCommand{\rev}[1]{#1}
\else
    \newcommand{\holger}[1]{[{\bf \color{orange} HC: #1}]}
    \newcommand{\wang}[1]{[{\bf \color{magenta} SW: #1}]}
    \newcommand{\lin}[1]{[{\bf \color{lime} YL: #1}]}
    \newcommand{\julian}[1]{[{\bf \color{cyan} JK: #1}]}
    \newcommand{\nan}[1]{[{\bf \color{yellow} LN: #1}]}
    \DeclareRobustCommand{\rev}[1]{{\color{blue}#1}}
\fi

\newcommand{\method}{RVLoss}

\def\pc{\mathbf{P}_}
\def\flow01{\mathbf{F}_{0\rightarrow1}}

\newcommand{\Eq}[1]{Eq.~(\ref{eq:#1})}

\newcommand{\fig}[1]{Fig.~\ref{fig:#1}}
\newcommand{\tab}[1]{Tab.~\ref{tab:#1}}

\newcommand{\sect}[1]{Sec.~\ref{sec:#1}}

\newcommand{\alg}[1]{Algorithm~\ref{alg:#1}}

\usepackage[ruled,vlined]{algorithm2e}

\usepackage{graphicx}
\usepackage{amsmath}
\usepackage{amssymb}
\usepackage{pifont}
\newcommand{\cmark}{{\color{ForestGreen} \ding{51}}}%
\newcommand{\xmark}{{\color{red} \ding{55}}}%
\usepackage{multirow}
\usepackage[table,svgnames]{xcolor}
\definecolor{lightgray}{gray}{0.92}
\definecolor{lightblue}{RGB}{220,235,250}
\usepackage{makecell}
\usepackage{caption}

\title{RVLoss: Runoff Vote Loss for Self-Supervised LiDAR Scene Flow Estimation}

\author{
Shiming Wang$^1$ \quad Liangliang Nan$^1$ \quad  Julian F.P. Kooij$^1$ \quad  Holger Caesar$^1$ \quad  Yancong Lin\textsuperscript{2,}\footnotemark[1] \\
$^1$TU Delft, the Netherlands \qquad  $^2$University of Nottingham, UK\\
}

\begin{document}

\maketitle

\begingroup
  \renewcommand{\thefootnote}{\fnsymbol{footnote}}
  \footnotetext[1]{Corresponding author.}
\endgroup

\begin{abstract}
LiDAR scene flow estimates point-wise motion between two consecutive scans, referred to as the source and target. 
Leading self-supervised methods typically minimize the Chamfer loss, the nearest-neighbor distance between the flow-compensated source and the target. 
However, nearest-neighbor search does not enforce motion rigidity, often leading to inconsistent flows within object instances. 
Existing approaches address this issue with additional regularization terms, but flow consistency among points remains limited, especially for large objects.
We propose \textbf{\method{}}, a self-supervised loss that incorporates motion rigidity by design through a \textit{runoff vote} mechanism. 
Our key observation is that the point-wise motion, calculated from nearest-neighbor search, can often be grouped into a small set of dominant flow candidates by voting (top-$k$ voting). 
Furthermore, when compensating the source by these candidates, the flow that best represents the underlying rigid motion often yields the highest consensus after a second voting (top-$1$ voting).
Based on this insight, we incorporate the two-stage \textit{runoff vote} into loss design and create cluster-wise rigid flows and free-form flows as pseudo-labels for self-supervised learning. 
\method{} can be seamlessly integrated into existing feedforward architectures. 
Experiments on the Argoverse2 2026 Scene Flow Challenge show that models trained with \method{} achieve state-of-the-art performance among self-supervised approaches, outperforming baseline models trained with alternative loss designs by 20\% and reducing the mean dynamic normalized end-point error from $0.202$ to $0.163$.
%
Moreover, cross-dataset evaluations demonstrate consistent performance improvements across four additional datasets. 
\rev{Code is available at \url{https://github.com/tudelft-iv/RVLoss}.}

\end{abstract}

\begin{figure}[t]
    \centering
    \includegraphics[width=\columnwidth]{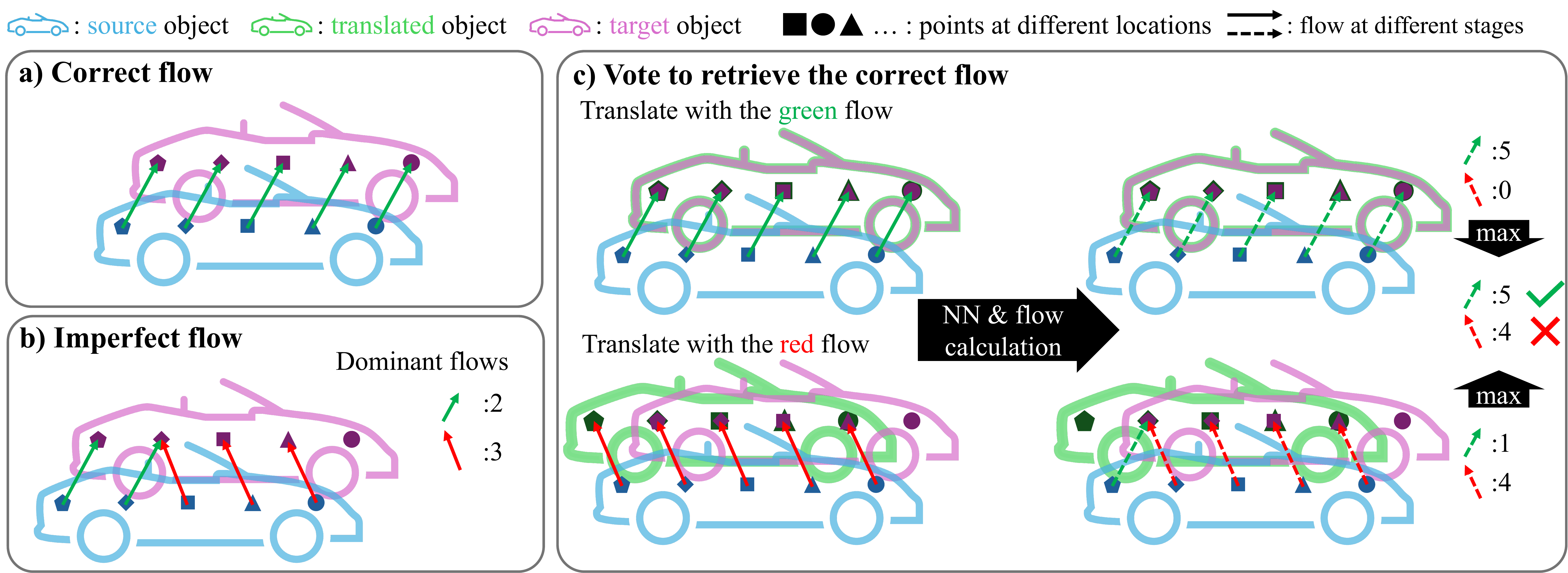}
    \caption{\textbf{Overview of a runoff vote.} 
    a) A car moves rigidly and thus exhibits consistent flows among points. 
    b) The widely used Chamfer distance breaks motion rigidity due to the local nearest-neighbor search. 
    As a result, multiple flow candidates emerge within the object. 
    Although the correct flow is among these candidates, it may not be the top-voted one ($1^{\text{st}}$ stage).
    c) We translate the source by those flow candidates and measure how well it aligns with the target by an additional voting ($2^{\text{nd}}$ stage). The flow receiving the highest number of votes is selected as the rigid flow, representing the underlying motion of the car.
    }
    \label{fig:intro}
\end{figure}

\section{Introduction}
\label{sec:intro}
LiDAR scene flow is the per-point 3D motion between two consecutive point clouds, and estimating scene flow is essential for many downstream tasks in autonomous driving, such as object discovery~\cite{najibi2022motion, lentsch2024union}, dynamic object reconstruction \cite{zhang2025himo, Hur2026UFO}, and BEV feature alignment \cite{wang2026asyncbev}.  
Self-supervised methods have recently gained increasing attention, as they reduce reliance on costly manual annotations.
Progress has been substantial, as demonstrated by the annual Argoverse2 Scene Flow Challenge~\cite{Argoverse2,khatri2024trackflow}, where leading methods achieve both high accuracy and real-time inference. 
The majority of leading works follow a feedforward design and use the Chamfer distance as the loss function, where the goal is to minimize the spatial distance between a point and its nearest neighbor across time after flow compensation~\cite{zhang2025seflow,lin2025voteflow, zhang2026teflow}.
However, nearest-neighbor search breaks motion rigidity (a common property for traffic participants in autonomous driving scenarios), as shown in \fig{intro}. As a result, there are often inconsistent flows, particularly for large objects such as cars, buses, and trucks.

A common strategy for enhancing motion rigidity is to introduce additional regularization terms. For instance, Zhang et al.~\cite{zhang2025seflow, zhang2025himo} apply regularization to static, dynamic, and clustered points, where point labels are precomputed using off-the-shelf algorithms.
Beyond explicit regularization, Lin et al.~\cite{lin2025voteflow} propose a differentiable module that enforces motion rigidity from an architectural perspective by aggregating features over predefined local neighborhoods to capture collective motion cues.
While these approaches improve performance, the resulting flow predictions still exhibit inconsistent motion within rigid objects. 

In this paper, we investigate how to incorporate motion rigidity directly into the loss design, eliminating the need for delicate architectural modification or multiple regularization terms. 
Our key observation is that point-wise motions of rigidly moving points, calculated from nearest-neighbor search, can often be grouped into a small set of dominant flows (top-$k$), and crucially, the ideal flow (top-$1$) that best characterizes the underlying rigid motion typically resides among these candidates, as shown in \fig{intro} b).
Intuitively, the top-$k$ candidates provide a limited number of viable solutions, which can be further evaluated one by one by shifting the source point cloud toward the target and by checking for consistency among clustered points, as illustrated in \fig{intro} c). 
%
Both the selection of top-$k$ candidates and the identification of the top-$1$ can be implemented via \textit{voting}, i.e., aggregating the total number of points that share the same motion. 
Based on this insight, we propose \textbf{\method{}}, a self-supervised loss function built on two-stage cascaded voting for scene flow estimation, also known as a \textit{runoff vote}. 
\method{} generates cluster-wise rigid flows and free-form flows, which are treated as pseudo-labels for self-supervised learning.

\method{} differs from directly selecting the top-$1$ dominant flow via a single-stage voting, which often underestimates the actual motion. 
For large objects (e.g., trucks) that exhibit significant spatial overlap after motion, nearest-neighbor search tends to favor correspondences with minimal displacement. Such matches, however, fail to capture the true object motion, especially on geometrically featureless surfaces.
Selecting the top-$k$ candidates instead accounts for multiple plausible displacements shared across points, yielding a compact set of hypotheses and enabling efficient downstream voting.


%
Our experiments on the Argoverse2 2026 Scene Flow Challenge\footnote{\href{https://www.argoverse.org/sceneflow.html}{https://www.argoverse.org/sceneflow.html}} demonstrate consistent performance improvement of \method{} over other loss designs. 
Cross-dataset evaluation confirms the strong generalization ability of models supervised by \method{}, yielding top performance on four additional datasets.
%
To summarize, our contributions are as follows:
\begin{itemize}
    \item We propose \textbf{\method{}} for self-supervised scene flow learning based on a \textit{\underline{R}unoff \underline{V}ote} process.
    As its name suggests, \method{} uses two-stage cascaded voting to identify cluster-wise rigid motions, and uses these as pseudo-labels for self-supervised learning. 
    This way, motion rigidity is encoded directly by the pseudo-labels, rather than as a learning objective of a separate loss term.
    
 %
    \item 
    \method{} is architecture-agnostic and can be seamlessly integrated with mainstream scene flow backbones.
    Models supervised by \method{} achieve leading performance on the Argoverse2 2026 Scene Flow Challenge~\cite{li2025uniflow} in both in-domain test and cross-domain evaluation across four other datasets, in particular on rigid moving object categories.
  %

    
    
\end{itemize}

 
\section{Related Work}
\label{sec:related_work}
Our work builds on four lines of research: efficient backbone design, self-supervised loss design, rigid motion modeling, and voting-based point cloud processing.

%
\textbf{Backbone design for LiDAR scene flow estimation.}
LiDAR scene flow methods commonly adapt feedforward architectures to predict per-point motion~\cite{behl2019pointflownet,liu2019flownet3d,puy20flot,gu2019hplflownet,liu2019meteornet,kittenplon2021flowstep3d,wang2021festa,cheng2022bi,wang2020flownet3d++}. However, early models mainly operate on downsampled point clouds, typically up to 8k points, and therefore struggle with full-scale LiDAR scans~\cite{wu2020pointpwc,kittenplon2021flowstep3d,jin2022deformation,li2022rigidflow}. 
FastFlow3D~\cite{jund2021scalable} addresses this limitation as the first feedforward model for real-time large-scale scene flow, and subsequent methods build on this design for further improvement~\cite{vedder2023zeroflow,zhang2024deflow,lin2025voteflow,zhang2025himo,khoche2025ssf}. 
Recent works extend feedforward estimation from frame pairs to LiDAR sequences~\cite{zhang2025deltaflow,kim2024flow4d}. 
In parallel, test-time optimization also demonstrates strong performance~\cite{li2021neural,li2023fast,hoffmann2025floxels,vedder2024eulerflow,lin2024icpflow}. However, the time-consuming inference limits practical deployment.
In this work, we build on established feedforward backbones~\cite{zhang2024deflow,zhang2025deltaflow} and focus on improving supervision through \method{}.

\textbf{Loss design for self-supervised scene flow learning.}
Early scene flow methods are predominantly supervised, requiring precise annotations from real-world datasets~\cite{behl2019pointflownet,liu2019flownet3d,huang2022dynamic,jin2022deformation} or synthetic data~\cite{Dosovitskiy_2015_ICCV}. 
To reduce reliance on manual annotation, self-supervised methods commonly adopt cycle-consistency-based objectives~\cite{baur2021slim,mittal2020just,wu2020pointpwc,li2021neural}. 
However, these objectives typically estimate per-point motion independently and lack an explicit object-level rigidity prior, leading to intra-object motion inconsistency under sparse LiDAR sampling. 
Recent works address this issue with additional regularizers~\cite{zhang2025seflow,zhang2025himo,hoffmann2025floxels}, while TeFlow~\cite{zhang2026teflow} strengthens supervision through multi-frame temporal consistency. 
In contrast, we directly incorporate motion rigidity into loss design through a \textit{runoff vote} mechanism, eliminating the need for additional regularization terms.

%
\textbf{Rigid motion priors.}
Rigid motion is a common property in autonomous driving and has been incorporated into scene flow estimation through architectural design and loss regularization. From the model-design perspective, ICP-Flow~\cite{lin2024icpflow} estimates rigid transformations of clustered points with ICP~\cite{besl1992method}. 
RigidFlow~\cite{li2022rigidflow} integrates ICP into a learning-based framework.
PointPWC~\cite{wu2020pointpwc} promotes local consistency through cost volumes and neighborhood feature aggregation. 
More recently, VoteFlow~\cite{lin2025voteflow} introduces a differentiable voting module to identify shared translations within clustered points. 
Unlike these methods that encode rigidity through architectural components, we enforce motion rigidity directly through loss design.
From the loss-design perspective, prior works typically encourage motion consistency through local regularization terms~\cite{vidanapathirana2024multi,zhang2025seflow,zhang2025himo,hoffmann2025floxels}. 
\rev{In contrast, \method{} uses a \textit{runoff vote} to encode the rigidity prior directly in the pseudo-labels used to train a scene-flow backbone, without modifying its architecture or adding a separate regularization term.}

%

\textbf{Voting mechanisms for point cloud processing.}
Voting is a classical technique for extracting geometric primitives~\cite{duda1972use, ballard1981generalizing,lin2022vpd,lin2020deep}. 
It has been widely used in point cloud processing, including segmentation~\cite{milletari2015robust}, detection~\cite{lehmann2011fast}, tracking~\cite{milletari2015universal}, and pose estimation~\cite{sun2010depth}, where sparse local evidence is aggregated into object-level hypotheses. 
Recent scene flow methods have also incorporated voting mechanisms. ICP-Flow~\cite{lin2024icpflow} uses voting to identify dominant translations, while VoteFlow~\cite{lin2025voteflow} extends voting into feature space.
%
\rev{Specifically, VoteFlow~\cite{lin2025voteflow} accumulates votes over displacement bins
according to the similarity of learned pillar features, and this voting
module remains part of the network at inference.}
%
\method{} instead incorporates
voting into the training loss: a first vote over nearest-neighbor
displacements proposes a small set of candidate translations, each
candidate guides a new geometric matching, and a second vote over the
matched displacements yields the supervision. Consequently, RVLoss
trains existing backbones without architectural changes.

\section{Methodology}
\label{sec:method}
This section describes the problem definition and our proposed \method{}: a self-supervised loss for feedforward scene flow estimation, compatible with state-of-the-art backbone designs. 

\begin{figure}[t]
    \centering
    \includegraphics[width=\columnwidth]{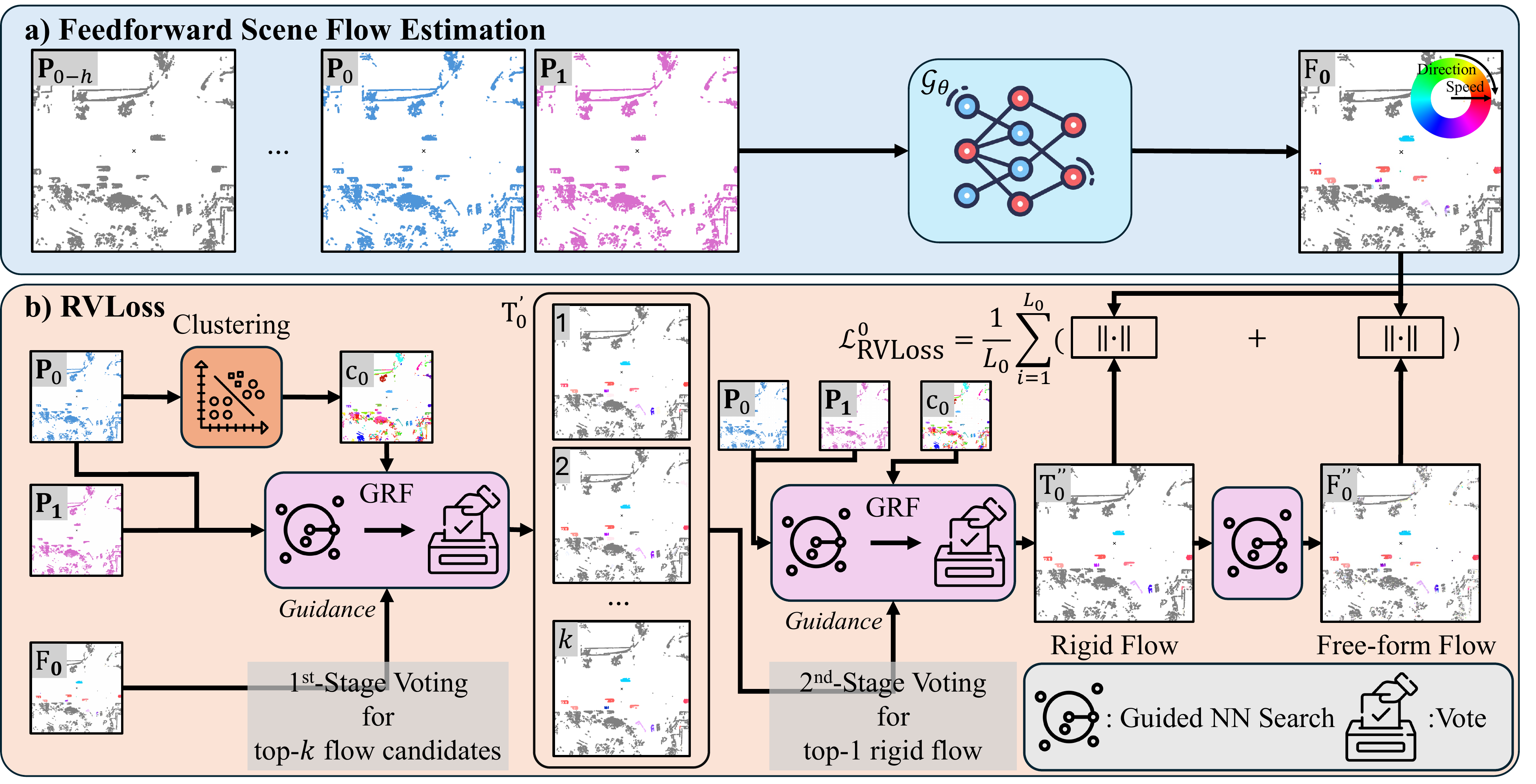}
    \caption{\textbf{Overview of \method{}.} 
a) A typical feedforward scene flow estimator $\mathcal{G}_{\theta}$ takes as input a sequence of LiDAR scans ($\geq 2$ frames), $\mathbf{P}_{0-h}, \dots, \mathbf{P}_{0}, \mathbf{P}_{1}$, and predicts per-point flows $\mathbf{F}_{0}$.
b) \method{} employs a \textit{runoff vote} mechanism to estimate rigid and free-form flows over clustered points during each forward pass, and uses the resulting flows as pseudo-labels. The \textit{runoff vote} consists of a two-stage voting procedure built upon a shared operation, termed \textit{Guided Rigid Flow Estimation (GRF)}.
Specifically, GRF first uses the predicted flow $\mathbf{F}_{0}$ as guidance to select the top-$k$ rigid flow candidates $\mathbf{T}^{'}_{0}$ (Stage~1). Subsequently, GRF takes these top-$k$ candidates as guidance and identifies the top-$1$ rigid flow $\mathbf{T}^{''}_{0}$ (Stage~2). In addition to rigid flow, \method{} also computes a free-form flow $\mathbf{F}^{''}_{0}$ as complementary supervision to relax the rigid-motion assumption and to mitigate the quantization error introduced by voting.
The objective is to minimize the residual between the predicted flow and two supervision flows: the top-$1$ rigid flow and the free-form flow.
    }
    \label{fig:main}
\end{figure}

\subsection{Problem Definition}
Given a pair of consecutive LiDAR point clouds, compensated by ego motion, $\mathbf{P}_{0} \in \mathbb{R}^{L_0 \times 3} = \{ \mathbf{p}_0^i \in \mathbb{R}^3 \}_i^{L_0} $ and $\mathbf{P}_{1} \in \mathbb{R}^{L_1 \times 3} = \{ \mathbf{p}_1^j \in \mathbb{R}^3 \}_j^{L_1}$, 
a neural network $\mathcal{G}_{\theta}$ produces a per-point flow field, $\mathbf{F}_{0} \in \mathbb{R}^{L_0 \times 3} = \{ \mathbf{f}_{0}^i \in \mathbb{R}^3 \}_i^{L_0}$, 
such that $\mathbf{P}_{1} \approx \mathbf{P}_{0} + \mathbf{F}_{0}$. $L_0$ and $L_1$ denote the number of points in $\mathbf{P_0}$ and $\mathbf{P_1}$.
Optionally, multiple historical frames $\mathbf{P}_{0-h}$, $\cdots$, $\mathbf{P}_{0}$ can be incorporated as input to improve performance, where $h$ is 
the number of historical frames~\cite{zhang2025deltaflow, lin2025voteflow}.
The training objective of the feedforward scene flow estimation is defined as:
\begin{align}
    \mathbf{F}_{0} =&\ \mathcal{G}_{\theta}\left(\mathbf{P}_{0-h}, \cdots ,\mathbf{P}_{0}, \mathbf{P}_{1}\right), 
    \\
    \theta^* =&\ \arg
\min_{\theta \in \Theta}
\;
\mathbb{E}_{(\pc0, \pc1) \sim \mathcal{D}}
\left[
\mathcal{L}_{\text{}} 
\left(\mathbf{P}_{0}, \mathbf{P}_{1},
\mathbf{F}_{0}
\right)
\right],
\end{align}
where $\mathcal{D}$ denotes the distribution of consecutive point cloud pairs, and $\mathcal{L}$ represents a self-supervised loss.
%
%
The bidirectional Chamfer loss $\mathcal{L}_{CD}$ is widely used in self-supervised scene flow estimation~\cite{fan2017point, li2021neural, li2023fast, zhang2025seflow, vedder2024eulerflow, lin2025voteflow, hoffmann2025floxels, zhang2026teflow}, i.e.:
\begin{align}
\label{eq:chamfer_distance}
\mathcal{L}_{\text{CD}}
=
\frac{1}{L_0}
\sum_{i=1}^{L_0}
\mathcal{L}_{\text{NN}}(\mathbf{p}_0^i + \mathbf{f}_{0}^i , \mathbf{P}_{1})
+
\frac{1}{L_1}
\sum_{j=1}^{L_1}
\mathcal{L}_{\text{NN}}(\mathbf{p}_1^j , \mathbf{P}_{0} + \mathbf{F}_{0}),
\end{align}
where $\mathcal{L}_{\text{NN}}$ calculates the distance between a point and its nearest neighbor in another point cloud, defined by $\mathcal{L}_{\text{NN}}(\mathbf{p}_{0}, \mathbf{P}_{1}) = \min_{\mathbf{p}_{1}^{j} \in \mathbf{P}_{1}}\| \mathbf{p}_{0} - \mathbf{p}_{1}^{j} \|$.


Despite its popularity, $\mathcal{L}_{\text{CD}}$ inherits a critical limitation from nearest-neighbor search: the lack of motion rigidity, which leads to inconsistent flow predictions.
To address this issue, recent works~\cite{zhang2025seflow, zhang2025himo} introduce multiple regularization terms to encourage coherent motion within each cluster. 
More importantly, adding extra regularization does not guarantee motion rigidity, demonstrating the need for an inherently motion-rigid loss function and thus motivating our \method{}.

\subsection{\method{}}\label{sec:loss}

\begin{figure}[t]
    \centering
    \includegraphics[width=\columnwidth]{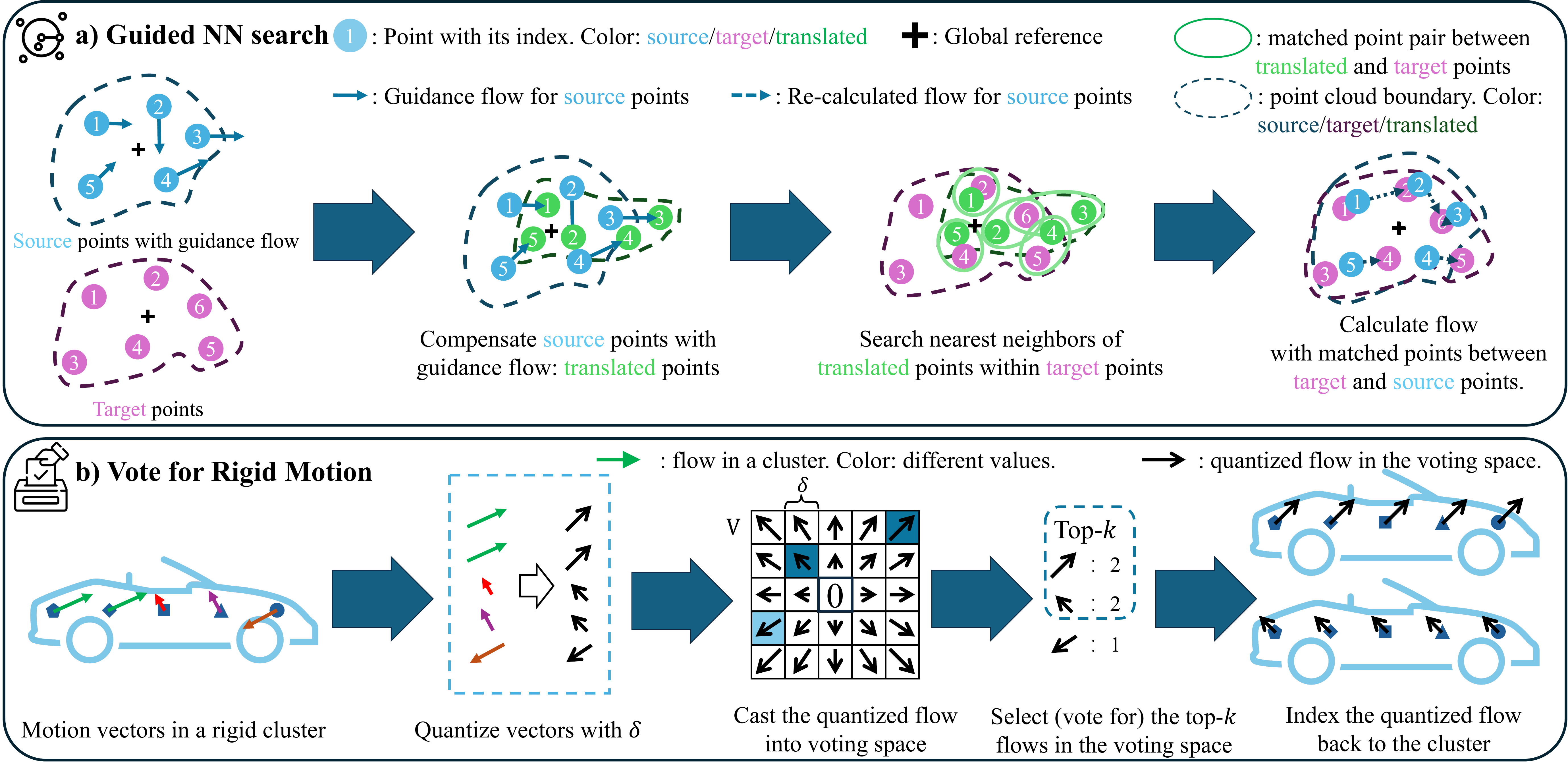}
    \caption{\textbf{Overview of GRF.} 
    \textbf{a) Guided NN Search} first translates the \textcolor{cyan}{source} points using the guidance flow and then finds their nearest neighbors in the \textcolor{HotPink}{target} point cloud. The point-wise motion between each source point and its matched target point is used for the subsequent processing.
    \textbf{b) Vote for Rigid Motion} quantizes the point-wise flows within each cluster and casts them into a predefined voting space. The top-$k$ flow candidates are then selected and assigned to all points within the cluster.
    }
    \label{fig:grf}
\end{figure}

Our proposed \method{} enables self-supervised learning by generating pseudo flow labels during training.
It aims to create increasingly more reliable rigid flow estimates during every forward pass, by (1) using the model's increasingly more accurate flow predictions as guidance to calculate the point-wise motion from observed data, and (2) pooling motions across pre-computed point clusters to estimate top-$k$ cluster-wise rigid flows via voting.
The clustering of points is done beforehand by HDBSCAN following common practice~\cite{zhang2025seflow, mcinnes2017hdbscan, campello2013hdbscan}.
We refer to this procedure as \textit{Guided Rigid Flow Estimation} or \textit{GRF} in short, as shown in \fig{grf}. A pseudo-implementation of GRF (\alg{grf}) is also available in \sect{algorim_grf}.

In practice, we apply GRF twice (see \fig{main}), resulting in a cascaded runoff vote process.
In the $1^{\text{st}}$ stage, GRF uses the model's predicted flow as guidance to generate the top-$k$ most likely rigid flow candidates per cluster (top-$k$ flows).
In the $2^{\text{nd}}$ stage, GRF uses each of the $k$ flow candidates as guidance, letting each point vote $k$ times to find consistent object motion across all points within the cluster. The flow with the highest vote count is deemed the rigid flow (top-$1$ flow).
We also add a free-form loss term to relax the rigid motion assumption and to handle the quantization errors in voting by performing an additional nearest-neighbor lookup using the top-$1$ rigid flow as guidance. 
With the rigid and free-form flows established, the forward pass is supervised by a distance-based loss, e.g., $\ell_1$ loss, between the predicted flow and pseudo flows.
The following subsections first describe GRF, and then elaborate on its use in both stages to create pseudo flows.

\subsubsection{Guided Rigid Flow Estimation (GRF)}
GRF takes as input the source $\mathbf{P} \in \mathbb{R}^{L_{p}\times 3}$, the target $\mathbf{Q} \in \mathbb{R}^{L_{q} \times 3}$, the cluster indices of the source $\mathbf{c} \in \mathbb{Z}^{L_{p}}$,
a guidance flow $\mathbf{F} \in \mathbb{R}^{L_{p} \times 3}$, 
and a hyperparameter $k \in \mathbb{Z}^{+}$ that determines the number of top-$k$ flow candidates per cluster.
GRF consists of the following two steps:

\textbf{Guided nearest-neighbor search.}
It checks where each point in $\mathbf{P}$ moves to according to the guidance flow $\mathbf{F}$, and identifies the nearest point in $\mathbf{Q}$. Then it computes point-wise flows $\mathbf{M} \in \mathbb{R}^{L_{p} \times 3} = \operatorname{NN}(\mathbf{P}+\mathbf{F}, \mathbf{Q}) - \mathbf{P}$, where $\operatorname{NN}$ indicates the nearest-neighbor search. The whole process is shown in \fig{grf} a).

\textbf{Vote for rigid motion.}
Points within each cluster cast a vote, and the top-$k$ flow candidates per cluster are returned, as shown in \fig{grf} b).
To be specific, we create a uniformly spaced 2D voting space over $x$ and $y$ dimensions\footnote{We follow the flat ground assumption and thus assume no motion along the $z$ axis.}, 
bounded by the minimal and maximal motion along each dimension, 
resulting in $\mathbf{V} \in \mathbb{Z}^{c \times m \times n}$, where $m \times n$ are the spatial dimensions of the voting space over $x$ and $y$, and $c$ is the total number of clusters.
In practice, we set the voting step $\delta$, i.e., the distance between two adjacent bins in the voting space, to be $1$ cm. 
For the $i^{th}$ motion vector $\mathbf{M}_{i} \in \mathbb{R}^{3}$, we cast a vote into the corresponding bin where it falls after quantization. All votes are equally weighted, i.e., each vote increases the bin count by $1$. 
After voting, we select the top $k$ (quantized) flows  $\mathbf{D} \in \mathbb{R}^{c \times k \times 3}$ by the number of votes for each cluster, after non-maximum suppression. 
We further index $\mathbf{D}$ using cluster indices $\mathbf{c}$ to obtain the point-wise flow candidates $\mathbf{T}\in \mathbb{R}^{L_{p} \times k \times 3}$.
For simplicity, we use $\mathbf{T} = \operatorname{GRF}(\mathbf{P}, \mathbf{Q}, \mathbf{F}, \mathbf{c}, k)$ to illustrate GRF.

\subsubsection{\texorpdfstring{$1^{\text{st}}$}{1st} stage voting for top-\texorpdfstring{$k$}{k} rigid flows}\label{sec:topk_voting}
The goal of this stage is to generate a set of top-$k$ flow candidates using GRF.
The input to GRF consists of the source $\mathbf{P}_{0}\in \mathbb{R}^{L_{0} \times 3}$, 
the target $\mathbf{P}_{1} \in \mathbb{R}^{L_{1} \times 3}$, 
cluster indices $\mathbf{c}_{0} \in \mathbb{Z}^{L_{0}}$, 
initial flow prediction $\mathbf{F}_{0}\in \mathbb{R}^{L_{0}\times 3}$ (the guidance flow) by a neural network $\mathcal{G}_{\theta}$, 
and the number of top-$k$ flow candidates.
The output of GRF is top-$k$ rigid flow candidates $\mathbf{T}_{0}^{'}\in \mathbb{R}^{L_{0} \times k \times 3}$, representing $k$ most likely rigid flows per cluster.
We represent the $1^{st}$ stage by $\mathbf{T}_{0}^{'} = \operatorname{GRF}(\mathbf{P}_{0}, \mathbf{P}_{1}, \mathbf{F}_{0},\mathbf{c}_{0}, k)$.

\subsubsection{\texorpdfstring{$2^{\text{nd}}$}{2nd} stage voting for top-1 rigid flow}
\label{sec:top1_voting}
The goal of this stage is to select, from the top-$k$ candidates (i.e., $\mathbf{T}_{0}^{'}\in \mathbb{R}^{L_{0} \times k \times 3}$), the one that best represents the collective motion of clustered points, resulting in $\mathbf{T}_{0}^{''}\in \mathbb{R}^{L_{0} \times 1 \times 3}$. 
$\mathbf{T}_{0}^{'}$ becomes the guidance flow in this stage and each source point votes $k$ times. 
This can be done efficiently by tensor broadcasting operations (see \alg{loss}).
For simplicity, we omit the details of broadcasting.
Thus, we can re-use GRF for the  $2^{\text{nd}}$ stage voting, parameterized by $\mathbf{T}_{0}^{''} = \operatorname{GRF}(\mathbf{P}_{0}, \mathbf{P}_{1}, \mathbf{T}_{0}^{'}, \mathbf{c}_{0},  1)$. 

\subsubsection{Loss formulation} 
\label{sec:pseudo_label}
With the obtained top-$1$ rigid flow, two loss terms can be computed to complete the forward pass.

\textbf{Rigid flow.} $\mathbf{T}_{0}^{''}$ is the quantized flow that indicates where clustered points move. 
We consider $\mathbf{T}_{0}^{''}$ as the rigid flow supervision.

\textbf{Free-form flow.} 
To avoid quantization error, 
we further conduct a guided nearest-neighbor search to retrieve the nearest neighbor of $\mathbf{P}_{0}$ from $\mathbf{P}_{1}$ under the guidance of $\mathbf{T}_{0}^{''}$.
As a result, we obtain the free-form point-wise flow $\mathbf{F}_{0}^{''} \in \mathbb{R}^{L_0 \times 3}$ 
depicting the motion from $\mathbf{P}_{0}$ to $\mathbf{P}_{1}$, 
i.e., $\mathbf{F}_{0}^{''} = \operatorname{NN}(\mathbf{P}_{0}+\mathbf{T}_{0}^{''}, \mathbf{P}_{1}) - \mathbf{P}_{0}$. 
We treat $\mathbf{F}_{0}^{''}$ as the free-form flow supervision. 

The total loss, which consists of rigid-flow and free-form-flow supervision, is defined as follows. The corresponding pseudo-implementation is provided in \sect{algorim_rvloss}.
\begin{align}
    \mathcal{L}_{\text{\method{}}}^{0} = 
    \frac{1}{L_0} \sum_{i=1}^{L_0} \big( 
    \underbrace{\|\mathbf{T}_{0}^{''} -\mathbf{F}_{0}\|}_{\text{Rigid flow}} + 
    \underbrace{\|\mathbf{F}_{0}^{''} -\mathbf{F}_{0}\|}_{\text{Free-form flow}} 
    \big).
\end{align}

\textbf{Bidirectional supervision.}
Similar to the bidirectional Chamfer loss (\Eq{chamfer_distance}), 
we also calculate the rigid and free-form flow from $\mathbf{P}_{1}$ to $\mathbf{P}_{0}$, $\mathbf{F}_{1}^{''}$ and $\mathbf{T}_{1}^{''}$ to ensure cycle-consistency.
To be specific, we first calculate the nearest-neighbor index of $ \mathbf{P}_{1}$ with respect to $\mathbf{P}_{0} + \mathbf{F}_{0}$, and then retrieve the predicted flow accordingly, denoted as $\mathbf{F}_{1}$, which characterizes the motion from $\mathbf{P}_{1}$ to $\mathbf{P}_{0}$.
Afterwards, 
we repeat the aforementioned $1^{\text{st}}$ stage voting and $2^{\text{nd}}$ stage voting procedures 
by swapping $\mathbf{P}_{1}$ and $\mathbf{P}_{0}$, and obtain $\mathbf{F}_{1}^{''}$ and $\mathbf{T}_{1}^{''}$.
When bidirectional optimization is adopted, the overall loss becomes:
$\mathcal{L}_\text{\method{}} = 
\frac{1}{L_0}
\sum_{i=1}^{L_0}\big(\|\mathbf{T}_{0}^{''} -\mathbf{F}_{0}\| + \|\mathbf{F_{0}^{''}} -\mathbf{F}_{0}\| \big) + 
\frac{1}{L_1} \sum_{j=1}^{L_1} \big(\|\mathbf{T}_{1}^{''} -\mathbf{F}_{1}\| + \|\mathbf{F_{1}^{''}} -\mathbf{F}_{1}\| \big).$
Within each training iteration, the resulting rigid and free-form flows are treated as fixed supervision, and no gradients are propagated through the clustering, nearest-neighbor assignment, or voting operations.


\section{Experiments}
\label{sec:exp}
We conduct benchmarking on the Argoverse2 (AV2) 2026 Scene Flow Challenge to demonstrate the advantage of \method{} over other loss designs. 

\subsection{Datasets and Evaluation Metrics}
\textbf{Datasets.} 
The AV2 2026 Scene Flow Challenge has been released to benchmark models across a variety of datasets. 
The key difference to previous challenges is the multi-dataset evaluation, composed of five mainstream autonomous driving datasets:
Argoverse2 (AV2) \cite{Argoverse2},  Waymo \cite{waymo, jund2021scalable}, nuScenes~\cite{caesar2020nuscenes}, TruckScenes~\cite{fent2024truckscenes} and AevaScenes~\cite{aevascenes}.
%

\textbf{Evaluation Metrics.}
Following the challenge protocol, we compare the Dynamic Normalized End-Point Error (EPE)~\cite{khatri2024trackflow} within a 35\,m radius around the ego vehicle. 
Unlike standard EPE, which calculates the L2 distance between predicted and ground-truth flow vectors, Dynamic Normalized EPE normalizes EPE by the mean speed within predefined motion buckets, providing a fairer comparison across multiple categories with different velocities. 
The four categories are: Car, Other Vehicle (O.V.), Pedestrian (Ped.), and Wheeled Vulnerable Road User (VRU).

\subsection{Implementation Details}
\textbf{Baselines.} 
We make comparisons with leading feedforward models, including SeFlow \cite{zhang2025seflow}, VoteFlow \cite{lin2025voteflow}, SeFlow++ \cite{zhang2025himo},  TeFlow \cite{zhang2026teflow}, SSF~\cite{khoche2025ssf}, and DeltaFlow~\cite{zhang2025deltaflow}. 
We take the numerical results directly from the leaderboard if provided; otherwise, we report results using the publicly released checkpoints by the authors.
The optimization-based models are currently absent from the leaderboard, and we are unable to test them due to high computational cost\footnote{The leading optimization-based model, EulerFlow~\cite{vedder2024eulerflow}, takes 24 hours on one NVIDIA V100 16GB GPU for a \textit{single} AV2 sequence.}.

%

\textbf{Implementation of our models.} 
We use \method{} to supervise the training of two scene flow backbones: multi-frame DeltaFlow~\cite{zhang2025deltaflow} and two-frame DeFlow \cite{zhang2024deflow}. We follow the official training schemes provided by OpenSceneFlow\footnote{\href{https://github.com/KTH-RPL/OpenSceneFlow}{https://github.com/KTH-RPL/OpenSceneFlow}}.
We take $k=10$ by default in the $1^{\text{st}}$ stage voting and discuss the selection of $k$ in \sect{ablation}.
We adopt the bidirectional optimization by default.
Training DeltaFlow+\method{} takes approximately 3 days on 4 NVIDIA A40 GPUs (48GB VRAM). 
%
%
\begin{table}[t]
\centering
\caption{In-domain performance on the AV2 \emph{test} split. \rev{Numerical results on scene flow prediction come from the AV2 2026 Scene Flow Challenge leaderboard.} 
`\#F\textsubscript{input}' denotes the number of input frames to the network. 
`\#F\textsubscript{loss}' denotes the number of input frames during loss calculation. 
RT stands for runtime per sequence (around 157 frames) in seconds, retrieved from TeFlow~\cite{zhang2026teflow}.
(\textbf{Best}, \underline{Second Best})
}
\label{tab:argo_test}
\setlength{\tabcolsep}{5pt}
\renewcommand{\arraystretch}{0.95}
\resizebox{\columnwidth}{!}{
\begin{tabular}{l l c c c | c c c c c | c c c c}
\toprule
\multirow{2}{*}{Backbone} &\multirow{2}{*}{Loss}& \multirow{2}{*}{\#F\textsubscript{input}} & \multirow{2}{*}{\#F\textsubscript{loss}} & \multirow[c]{2}{*}{RT (s)} & \multicolumn{5}{c|}{Dynamic Normalized EPE $\downarrow$} & \multicolumn{4}{c}{Three-way EPE (cm) $\downarrow$}   \\
\cmidrule(lr){6-10} \cmidrule(lr){11-14}
& & & &  & Mean & Car & O.V. & Ped. & VRU & Mean & FD & FS & BS \\
\midrule
\multicolumn{5}{l|}{\textit{Supervised methods}} \\
DeltaFlow & GT & 5 & - & 8.0 & 0.103 & 0.083 & 0.132 & 0.122 & 0.075 & 1.76 & 3.16 & 1.31 & 0.81  \\
SSF & GT & 2 & - & - & 0.160 & 0.115 & 0.170 & 0.217 & 0.137 & 2.36 & 4.71 & 1.69 & 0.67\\
\midrule
\multicolumn{5}{l|}{\textit{Self-supervised methods}} \\
DeFlow & SeFlow  & 2 & 2 & 7.2 & 0.313 & 0.236 & 0.295 & 0.441 & 0.273 & 4.76 & 10.93 & 2.23 & \underline{1.12} \\
VoteFlow  & SeFlow  & 2 & 2 & 13.0 & 0.290 & 0.226 & 0.301 & 0.383 & 0.252 & 4.52 & 10.36 & \underline{2.09} & \underline{1.12}  \\
DeFlow++ & SeFlow++  & 3 & 3 & 10.0 & 0.283 & 0.202 & 0.308 & 0.387 & 0.234 & 4.20 & 9.49 & \textbf{2.02} & \textbf{1.08}  \\
\rowcolor{lightblue}
DeFlow  & \method{} (ours) & 2 & 2 & 7.2 & 0.256 & \underline{0.160} & 0.254 & 0.391 & 0.218 & 3.63 & 7.14 & 2.16 & 1.60 \\
DeltaFlow  & TeFlow  & 5 & 5 & 8.0 & \underline{0.202} & 0.167 & \underline{0.221} & \underline{0.273} & \underline{0.146} & \underline{3.54} & \underline{7.02} & 2.18 & 1.42 \\

\rowcolor{lightblue}
DeltaFlow & \method{} (ours) & 5 & 2 & 8.0  & \textbf{0.163} & \textbf{0.116} & \textbf{0.154} & \textbf{0.242} & \textbf{0.141} & \textbf{2.95} & \textbf{4.92} & 2.12 & 1.82 \\
\bottomrule
\end{tabular}
}
\end{table}


\begin{figure}[htbp]
    \centering
    \includegraphics[width=\columnwidth]{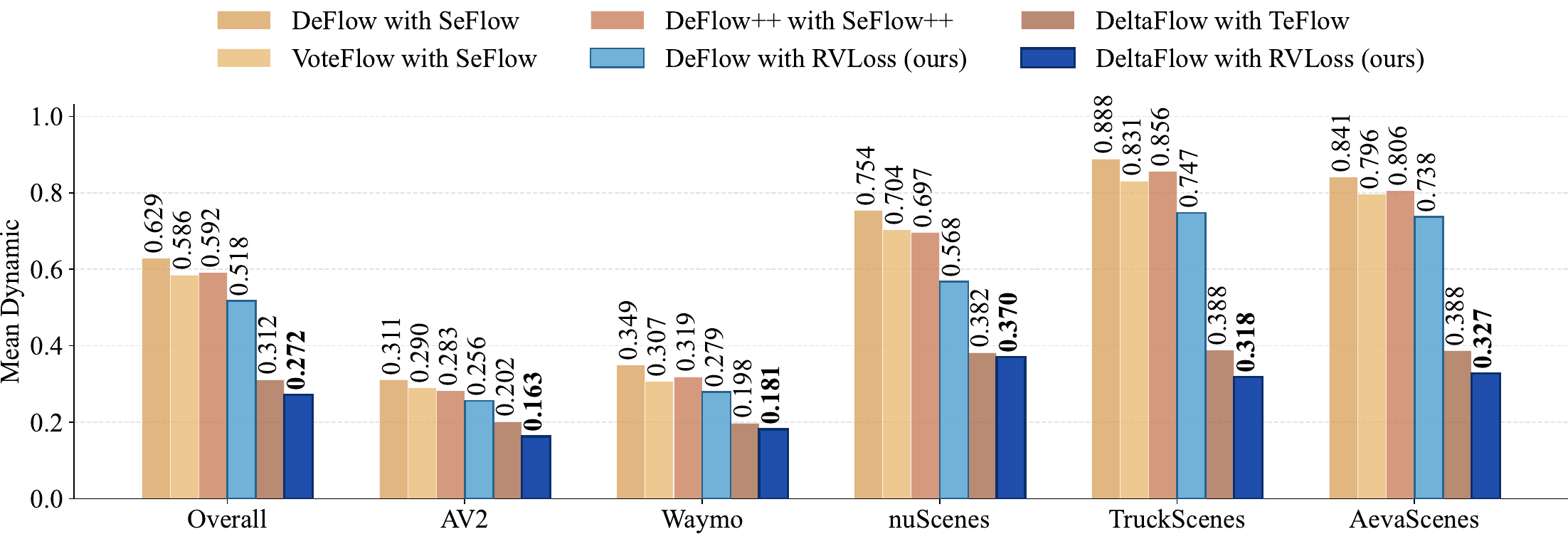}
    \caption{\textbf{Cross-domain evaluation performance.} We report the mean dynamic normalized EPE (Mean Dynamic) on multiple datasets used in the AV2 2026 Scene Flow Challenge \cite{li2025uniflow}. 
    DeltaFlow+\method{} demonstrates consistent performance improvement over all baselines.
    }
    \label{fig:cross_data_test}
\end{figure}

\subsection{Quantitative Comparison}

\textbf{In-domain evaluation.}
\tab{argo_test} demonstrates the in-domain test on the AV2 \textit{test} split, which is part of the AV2 2026 Scene Flow Challenge. All models are trained on the AV2 \textit{train} split. We refer to all models by \textsc{backbone + loss}.
DeltaFlow+\method{} achieves the best performance among all self-supervised models when comparing mean dynamic normalized EPE.  
Particularly, given the DeltaFlow backbone, \method{} outperforms the TeFlow loss on Car and Other Vehicle by approximately 5 p.p. (percentage points) and 7 p.p., demonstrating the advantage of \method{} over other self-supervised loss functions.
Similarly, given the DeFlow~\cite{zhang2024deflow} backbone, \method{} is able to deliver better results than the SeFlow~\cite{zhang2025seflow} loss, verifying the effectiveness of the \textit{runoff vote} design.
Despite substantial improvement on Car and Other Vehicle, the gain on Wheeled VRU is less prominent, as the rigid motion assumption no longer stands and point density decreases.
When adopting \method{} during training, the DeltaFlow backbone reduces the mean dynamic normalized EPE by a large margin (up to 9 p.p.) over the DeFlow backbone, demonstrating the benefit of using temporal information. 
Although the performance gain induced by \method{} is encouraging, the gap to fully supervised baselines, e.g., DeltaFlow+GT, remains, particularly on Pedestrian and Wheeled VRU.
We also perform in-domain evaluation on Waymo \cite{jund2021scalable} and nuScenes \cite{caesar2020nuscenes} in \sect{in-domain_waymo_nusc}.

\begin{figure}[htbp]
    \centering
    \includegraphics[width=\columnwidth]{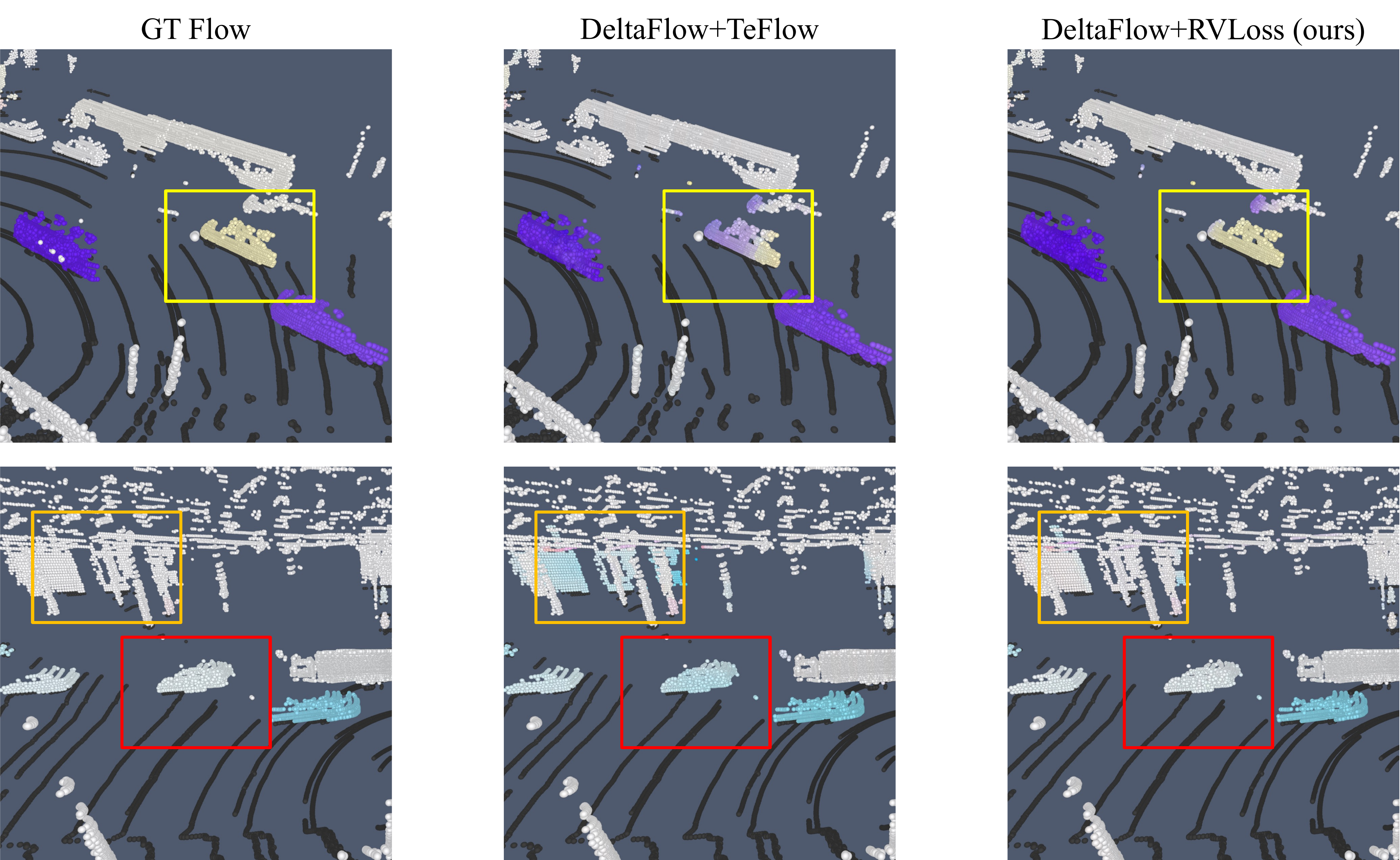}
    \caption{\textbf{Qualitative comparison}. Visualization of scene flow predictions from the DeltaFlow model supervised by the TeFlow loss and the proposed RVLoss. Point colors denote flow direction, and color saturation indicates flow magnitude. Objects are expected to exhibit coherent color patterns. Compared with the TeFlow~\cite{zhang2026teflow} loss, \method{} produces more uniform coloring within each object (highlighted in boxes), demonstrating more coherent scene flow estimation.
    }
    \label{fig:qualitative}
\end{figure}

\textbf{Cross-domain evaluation.}
We make further comparisons on multiple datasets adopted in the AV2 2026 Scene Flow Challenge, as shown in \fig{cross_data_test}. 
The models in this comparison are identical to those in \tab{argo_test}, i.e., all are trained on the AV2 \textit{train} split. DeltaFlow+\method{} demonstrates consistent performance improvement over the others on all five datasets. 
Notably, the error on nuScenes is nearly twice as high as on AV2 and Waymo. We speculate that the reduced point density is the main cause, as nuScenes uses a \textit{single} 32-beam LiDAR, resulting in 2$\times$ fewer beams than the other two.
The TruckScenes dataset differs from AV2, Waymo, and nuScenes, as it uses a truck as the ego vehicle and collects data mainly on highways. 
Despite the use of multiple 64-beam LiDAR sensors, the error remains high, suggesting that models trained on urban-centric and car-based datasets need further fine-tuning to better adapt to truck views and highway scenarios.
The AevaScenes dataset uses high-resolution FMCW LiDARs that differ substantially from rotating LiDAR sensors in the other four datasets. All models exhibit relatively high errors on AevaScenes, indicating a notable domain gap.
%
%
Overall, DeltaFlow+\method{} delivers not only the best result in the in-domain test but also in the cross-domain test, validating its generalization ability over other variants.

\begin{table}[tbp]
\centering
\caption{Ablation of each loss component on the AV2 \textit{val} split \cite{Argoverse2}. (\textbf{Best}, \underline{Second Best})}
\label{tab:ablation_loss}
\setlength{\tabcolsep}{5pt}
\renewcommand{\arraystretch}{0.95}
\resizebox{\columnwidth}{!}
{
\begin{tabular}{c c c|c c cc c | c c c c}
\toprule
\multirow{2}{*}{Free-form Loss} & \multirow{2}{*}{Rigid Loss} & \multirow{2}{*}{Bidirect Optim}  & \multicolumn{5}{c|}{Dynamic Normalized EPE $\downarrow$} &\multicolumn{4}{c}{Three-way EPE (cm) $\downarrow$} \\
\cmidrule(lr){4-8} \cmidrule(lr){9-12}
& & &  Mean & Car & O.V. & Ped. & VRU & Mean & FD & FS & BS \\
\midrule
\cmark & \cmark & \xmark & 0.284 & \underline{0.129} & 0.267 & \underline{0.442} & 0.299 & \underline{3.59} & \underline{7.31} & \underline{1.99} & \underline{1.48} \\

\xmark & \cmark & \cmark & 0.338 & 0.138 & 0.268 & 0.604 &0.346 & 3.70 & 8.27 & \textbf{1.69} & \textbf{1.14}\\
\cmark & \xmark & \cmark & \underline{0.275} & 0.139 & \underline{0.266} & \textbf{0.405} & \underline{0.288} & 4.68 & 7.48 & 3.62 & 2.94\\

\cmark & \cmark & \cmark & \textbf{0.262} & \textbf{0.121} & \textbf{0.246} & \textbf{0.405} & \textbf{0.277} & \textbf{3.47} & \textbf{6.84} & 2.05 & 1.53 \\
\bottomrule
\end{tabular}
}
\end{table}
\subsection{Ablation Study on Loss Composition}
We validate our design choices in \tab{ablation_loss}, using the DeFlow backbone~\cite{zhang2024deflow} as it takes less training time than DeltaFlow.
The comparison between the first and last rows shows that bidirectional optimization improves performance by enforcing cycle consistency.
While the free-form flow alone achieves the second-best performance on mean dynamic normalized EPE, it performs worst on static background regions, as the model tends to overestimate motion and predict most points as moving. Adding the rigid flow loss effectively reduces errors on static points.
Therefore, we use both rigid and free-form flows, together with bidirectional optimization, as our default configuration.

\begin{table}[!b]
\centering
\caption{\protect\rev{\textbf{Clustering sensitivity.}
Left: mean positive-cluster counts (and proportions) per category per frame on the
AV2 \textit{train} set.
Right: Dynamic Normalized EPE and Three-way EPE (cm).
(\textbf{Best}, \underline{Second Best}); \emph{Oracle} and \emph{Oracle FG}
are excluded from ranking. BG: background.}}
\label{tab:clustering-combined}
\label{tab:clustering-sensitivity}
\label{tab:cluster-counts-five}
\footnotesize
\setlength{\tabcolsep}{1.5pt}
\def\RVClusterCount#1#2{\begin{tabular}[t]{@{}r@{}}#1\\[-2pt]{\scriptsize(#2\%)}\end{tabular}}
\def\RVClusteringRows{%
\toprule
Clustering & \multicolumn{6}{c|}{Positive clusters per frame}
 & \multicolumn{5}{c}{Dynamic Normalized EPE $\downarrow$}
 & \multicolumn{4}{c}{Three-way EPE (cm) $\downarrow$} \\
\cmidrule(lr){2-7}\cmidrule(lr){8-12}\cmidrule(lr){13-16}
 & BG & Car & O.V. & Ped. & VRU & Total
 & Mean & Car & O.V. & Ped. & VRU & Mean & FD & FS & BS \\
\midrule
\emph{Under} & \RVClusterCount{68.22}{80.1} & \RVClusterCount{13.04}{15.3} & \RVClusterCount{1.51}{1.8} & \RVClusterCount{2.10}{2.5} & \RVClusterCount{0.34}{0.4} & 85.22 & 0.288 & 0.128 & 0.257 & 0.477 & 0.290 & \underline{3.57} & 7.32 & \textbf{1.93} & \textbf{1.46} \\
\emph{Default} & \RVClusterCount{108.24}{84.0} & \RVClusterCount{14.97}{11.6} & \RVClusterCount{1.97}{1.5} & \RVClusterCount{3.21}{2.5} & \RVClusterCount{0.46}{0.4} & 128.85 & \underline{0.262} & \textbf{0.121} & \underline{0.246} & \underline{0.405} & \underline{0.277} & \textbf{3.47} & \textbf{6.84} & \underline{2.05} & \underline{1.53} \\
\emph{Over} & \RVClusterCount{188.47}{86.9} & \RVClusterCount{20.18}{9.3} & \RVClusterCount{2.95}{1.4} & \RVClusterCount{4.46}{2.1} & \RVClusterCount{0.78}{0.4} & 216.85 & \textbf{0.256} & \underline{0.122} & \textbf{0.236} & \textbf{0.395} & \textbf{0.272} & 3.59 & \underline{6.85} & 2.22 & 1.71 \\
\midrule
\emph{Oracle} & \RVClusterCount{0.12}{0.4} & \RVClusterCount{18.42}{60.2} & \RVClusterCount{1.64}{5.4} & \RVClusterCount{8.49}{27.7} & \RVClusterCount{1.94}{6.3} & 30.61 & 0.299 & 0.132 & 0.280 & 0.471 & 0.314 & 3.40 & 7.43 & 1.79 & 0.99 \\
\emph{Oracle FG} & \RVClusterCount{115.89}{74.0} & \RVClusterCount{25.88}{16.5} & \RVClusterCount{2.70}{1.7} & \RVClusterCount{10.03}{6.4} & \RVClusterCount{2.07}{1.3} & 156.57 & 0.257 & 0.119 & 0.253 & 0.377 & 0.280 & 3.50 & 6.69 & 2.16 & 1.65 \\
\bottomrule%
}
\setbox0=\hbox{\rev{\begin{tabular}{@{}lrrrrrr|rrrrrrrrr@{}}\RVClusteringRows\end{tabular}}}%
\ifdim\wd0>\linewidth
  \PackageError{clustering-preview}{Horizontal clustering table exceeds text width}{}%
\fi
\rev{\begin{tabular*}{\linewidth}{@{\extracolsep{\fill}}lrrrrrr|rrrrrrrrr@{}}
\RVClusteringRows
\end{tabular*}}
\end{table}

\subsection{\rev{Clustering Sensitivity}}
\label{sec:clustering-sensitivity}

\begin{figure}[!t]
\centering
\includegraphics[width=\linewidth]{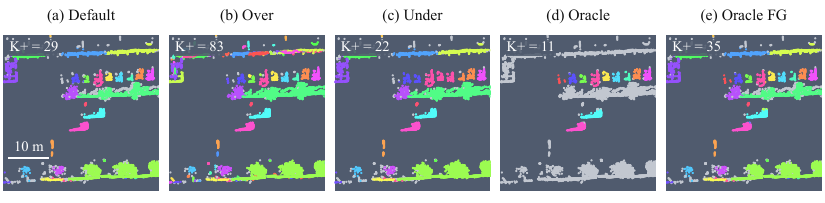}
\caption{\textbf{Qualitative comparison of five clustering settings.}
All panels show the same $40\times40$\,m point-cloud crop.
Colors distinguish positive-ID clusters within each panel;
gray indicates points with label~0.
$K_+$ counts the positive-ID clusters visible in the crop.}
\label{fig:clustering-sensitivity}
\end{figure}

\rev{
\method{} constructs supervision from precomputed point clusters.
To examine how clustering affects performance,
we compare five clustering settings using DeFlow \cite{zhang2024deflow} on
the AV2 \emph{val} split.
Following prior work~\cite{zhang2025seflow,zhang2025himo},
\emph{Default} uses HDBSCAN with minimum cluster size and minimum
samples both set to 20, a cluster-selection threshold of
$0.7$\,m, and $\alpha=1.1$.
\emph{Over} uses DBSCAN with radius $0.7$\,m and minimum samples 5.
\emph{Under} changes only \emph{Default}'s selection threshold
to $1.2$\,m.
\emph{Over} and \emph{Under} result in over-segmentation and under-segmentation
overall, respectively.}
We additionally consider two oracle clustering settings based on
the GT cuboid annotations.
%
%
\emph{Oracle} groups foreground points by GT cuboid instance and
assigns all remaining non-ground points to a single background
cluster with label~0.
For a closer comparison with \emph{Default}, \emph{Oracle FG}
preserves the same GT foreground clusters but uses \emph{Default}'s
local clusters outside the cuboids.
%
Each foreground or background cluster receives a unique positive ID
within the frame, while unclustered points retain label~0.
Fig.~\ref{fig:clustering-sensitivity} visualizes all five settings,
and Tab.~\ref{tab:clustering-combined} reports
mean per-category cluster counts per frame.

Tab.~\ref{tab:clustering-combined} shows that \emph{Over}
increases positive-cluster counts across all reported categories
while maintaining a mean dynamic normalized EPE comparable to \emph{Default}
($0.262\rightarrow0.256$).
\emph{Under} produces fewer clusters but increases mean dynamic normalized EPE
to $0.288$, with pedestrians exhibiting the largest degradation.
Finer clusters may better satisfy \method{}'s translation
approximation for dynamic objects: rigid rotation produces
spatially varying displacements that can be approximated by
piecewise translations.
This interpretation is consistent with \emph{Over}'s
improvements on Other Vehicle.

For static regions, 
larger clusters provide more votes and may
reduce sensitivity to LiDAR sampling noise, partial visibility,
and occlusion, whereas small fragments may yield less stable
estimates.
This is consistent with \emph{Under} achieving lower EPE-BS
and EPE-FS than \emph{Default} and \emph{Over}.
Similarly, \emph{Oracle} achieves
the lowest EPE-BS and EPE-FS, but its performance on the foreground is the worst.

Notably, \emph{Default} achieves a comparable mean dynamic normalized EPE to \emph{Oracle FG} ($0.262$ vs.\ $0.257$), suggesting
that \method{} does not require exact object-level foreground clustering.
This result demonstrates that, despite variations in performance, \method{} does not depend on a specific clustering configuration and is generally robust to over-segmentation.

\section{Conclusions}
\label{sec:conclusions}

We present \textbf{\method{}}, a self-supervised loss function for LiDAR scene flow estimation using a \textit{runoff vote} mechanism. 
Unlike the widely used Chamfer loss, \method{} explicitly incorporates the rigid-motion prior into the design without demanding a separate regularization term. 
Specifically, \method{} estimates cluster-wise rigid flows and point-wise free-form flows via a \textit{runoff vote} and uses them as pseudo-labels for self-supervised learning.
Extensive experiments on the AV2 2026 Scene Flow Challenge demonstrate that \method{} consistently improves flow estimation in both in-domain and cross-domain evaluations. 
For in-domain evaluation, DeltaFlow+\method{} achieves state-of-the-art performance on the AV2 \textit{test} set, reaching a mean dynamic normalized EPE of $0.163$ and substantially outperforming DeltaFlow+TeFlow ($0.202$). For cross-domain evaluation, DeltaFlow+\method{} achieves an average mean dynamic normalized EPE of $0.272$ across five datasets, outperforming DeltaFlow+TeFlow by $4$ percentage points. 
The advantage on Car and Other Vehicle validates the importance of a rigid-motion prior in loss design and demonstrates the effectiveness of the \textit{runoff vote} mechanism.
%
%

\textbf{Limitations and Future Work.}
%
Both voting stages in \method{} require precomputed cluster labels.
Future work could infer voting groups directly from predicted flow.
Another limitation is that the rigid-motion assumption generally holds for large traffic participants, such as cars, buses, and trucks, but may not hold for non-rigid or articulated agents, such as pedestrians and other vulnerable road users. Extending the motion prior beyond strict rigidity, e.g., toward part-level or deformable motion consistency, is an important direction for future work.
%
\section*{\rev{Acknowledgments}}
\rev{This work was supported by the 3D Urban Understanding (3DUU) Lab funded by the TU Delft AI Initiative.}
\bibliographystyle{unsrtnat}
\bibliography{main}

\newpage
\appendix
\section{Technical appendices and supplementary material}





\subsection{Pseudo-implementation of Guided Rigid Flow Estimation}
\label{sec:algorim_grf}
\begin{algorithm}[ht]
\caption{Guided Rigid Flow Estimation (PyTorch-style~\cite{paszke2019pytorch})}
\label{alg:grf}

\KwIn{$\mathbf{P} \in \mathbb{R}^{L_{P} \times 3}$, $\mathbf{Q} \in \mathbb{R}^{L_{Q} \times 3}$, $\mathbf{F} \in \mathbb{R}^{L_{P} \times 3}$, clusters $\mathbf{c} \in \mathbb{Z}^{L_{P}}$, $\delta\in \mathbb{R}^{+}$, $k\in \mathbb{Z}^{+}$}

\KwOut{Top-$k$ rigid flow per point $\mathbf{T} \in \mathbb{R}^{L_{P} \times k \times 3} $}

\noindent \colorbox{gray!20}{Guided nearest-neighbor search}

$\mathbf{M}\in \mathbb{R}^{L_{P} \times 3} \gets \texttt{NN}(\mathbf{P} + \mathbf{F}, \mathbf{Q}) - \mathbf{P}$ \tcp*{Point-wise motion}

\noindent \colorbox{gray!20}{Vote for top-$k$ rigid motion}

$c \gets \mathbf{c}.\texttt{max}() + 1$  \tcp*{Number of clusters}

$\mathbf{b}_{x} \in \mathbb{R}^{m} \gets \texttt{torch.arange}(\mathbf{M}[:,0].\texttt{min}(), \mathbf{M}[:,0].\texttt{max}(), \delta)$ \tcp*{Bins over $x$}

$\mathbf{b}_{y} \in \mathbb{R}^{n} \gets \texttt{torch.arange}(\mathbf{M}[:,1].\texttt{min}(), \mathbf{M}[:,1].\texttt{max}(), \delta)$ \tcp*{Bins over $y$}

$\mathbf{V} \in \mathbb{R}^{c \times m \times n} \gets \texttt{torch.zeros}(c, m, n)$ \tcp{Voting space over $x$ and $y$}

$\mathbf{v}_x \in \mathbb{Z}^{L_{P}} \gets \texttt{torch.bucketize}(\mathbf{M}[:,0], \mathbf{b}_{x})$ \tcp*{Flow indices after binning}

$\mathbf{v}_y \in \mathbb{Z}^{L_{P}} \gets \texttt{torch.bucketize}(\mathbf{M}[:,1], \mathbf{b}_{y})$ \tcp*{Flow indices after binning}

$\mathbf{V}[\mathbf{c}, \mathbf{v}_{x}, \mathbf{v}_{y}] \mathbin{+=} 1$
\tcp*{Majority voting}

$\mathbf{V} \gets \texttt{F.max\_pool2d}\left( \mathbf{V}.\texttt{unsqueeze}(0), 3, 1, 1 \right).\texttt{squeeze}(0)$  \tcp*{Non-max suppression}

$\mathbf{I} \in \mathbb{Z}^{c \times k} \gets \texttt{torch.topk}\left( \mathbf{V}.\texttt{view}(c, -1), k, -1 \right)$  \tcp*{Top-$k$ indices per cluster}

$\mathbf{D}_{x} \in \mathbb{R}^{c \times k} \gets \mathbf{b}_{x}[\mathbf{I} \mathbin{\%} m]$  \tcp*{Rigid flow over $x$ }

$\mathbf{D}_{y} \in \mathbb{R}^{c \times k} \gets \mathbf{b}_{y}[\mathbf{I} \mathbin{//} m \mathbin{\%} n]$  \tcp*{Rigid flow  over $y$ }

$\mathbf{D}_{z} \in \mathbb{R}^{c \times k} \gets \texttt{torch.zeros}(c, k)$  \tcp*{Zero flow over $z$ }

$\mathbf{D} \in \mathbb{R}^{c \times k \times 3}  \gets \texttt{torch.stack}([\mathbf{D}_{x}, \mathbf{D}_{y}, \mathbf{D}_{z}], -1)$  \tcp*{Top-$k$ rigid flow per cluster}

$\mathbf{T} \in \mathbb{R}^{L_{P} \times k \times 3}  \gets \mathbf{D}[\mathbf{c}]$ \tcp*{Top-$k$ rigid flow per point}

\Return{$\mathbf{T}$}\;
\end{algorithm}

\subsection{Pseudo-implementation of \method{}}
\label{sec:algorim_rvloss}

\begin{algorithm}[ht]
\caption{\method }
\label{alg:loss}

\KwIn{$\mathbf{P}_{0} \in \mathbb{R}^{L_{0} \times 3}$, $\mathbf{P}_{1} \in \mathbb{R}^{L_{1} \times 3}$, $\mathbf{F}_{0} \in \mathbb{R}^{L_{0} \times 3}$, $k\in \mathbb{Z}$, $\delta\in \mathbb{R}$}

\KwOut{$\mathbf{T_{0}^{''}}\in \mathbb{R}^{L_{0} \times 3}$, $\mathbf{F_{0}^{''}}\in \mathbb{R}^{L_{0} \times 3}$}

\noindent \colorbox{gray!20}{Clustering}

$\mathbf{c}_{0}\in \mathbb{Z}^{L_{0} } \gets \texttt{HDBSCAN}(\mathbf{P}_{0})$ \tcp*{Point clustering}

\noindent \colorbox{gray!20}{Top-$k$ rigid flows ($1^{\text{st}}$ stage)}

$\mathbf{T}_{0}^{'}\in \mathbb{R}^{L_{0} \times k \times 3} \gets \operatorname{GRF}(\mathbf{P}_{0}, \mathbf{P}_{1}, \mathbf{F}_{0}, \mathbf{c}_{0}, \delta, k)$ 
\tcp*{Top-$k$ rigid flows}

\noindent \colorbox{gray!20}{Top-$1$ rigid flow  ($2^{\text{nd}}$ stage)}

$\mathbf{P}_{0}^{'}\in \mathbb{R}^{L_{0} k \times 3} \gets \mathbf{P}_{0}[:, \texttt{None}, :].\operatorname{expand}(L_{0}, k, 3).\operatorname{view}(L_{0} \times k, 3)$
\tcp*{Broadcasting}

$\mathbf{c}_{0}^{'}\in \mathbb{R}^{L_{0} k} \gets \mathbf{c}_{0}.\operatorname{repeat\_interleave}(k)$
\tcp*{Broadcasting}

$\mathbf{T}_{0}^{'}\in \mathbb{R}^{L_{0} k \times 3} \gets \mathbf{T}_{0}^{'}.\operatorname{view}(L_{0} \times k, 3)$
\tcp*{Reshaping}

$\mathbf{T}_{0}^{''}\in \mathbb{R}^{L_{0} \times 1 \times 3} \gets \operatorname{GRF}(\mathbf{P}_{0}^{'}, \mathbf{P}_{1}, \mathbf{T}_{0}^{'}, \mathbf{c}_{0}^{'}, \delta, 1)$ 
\tcp*{Top-$1$ rigid flow}



$\mathbf{F_{0}^{''}}\in \mathbb{R}^{L_{0}\times 3} \gets \texttt{NN}(\mathbf{P}_{0} + \mathbf{T}_{0}^{''}.\operatorname{squeeze(1)}, \mathbf{P}_{1})  - \mathbf{P}_{0} $ \tcp*{ Free-form flow } 


\Return{$\mathbf{T}_0^{''},\mathbf{F}_0^{''}$}\;
\end{algorithm}

\subsection{In-domain evaluation on nuScenes and Waymo}
\label{sec:in-domain_waymo_nusc}
\begin{table}[htbp]
\centering
\caption{Dynamic Normalized EPE ($\downarrow$) on the nuScenes \cite{caesar2020nuscenes} \textit{val} set and the Waymo \cite{jund2021scalable} \textit{valid} set. (\textbf{Best}, \underline{Second Best})
}
\label{tab:nusc_waymo_val}
\setlength{\tabcolsep}{5pt}
\renewcommand{\arraystretch}{0.95}
\resizebox{\columnwidth}{!}
{
\begin{tabular}{l c|c c cc c | c c c c}
\toprule
\multirow{2}{*}{Method} & \multirow{2}{*}{Loss}   & \multicolumn{5}{c|}{ nuScenes \cite{caesar2020nuscenes} } &\multicolumn{4}{c}{Waymo \cite{jund2021scalable}} \\
\cmidrule(lr){3-7} \cmidrule(lr){8-11}
& &  Mean & Car & O.V. & Ped. & VRU & Mean & Car & Ped. & VRU \\
\midrule
DeFlow & SeFlow & 0.544 & 0.396 & 0.653 & 0.726 & 0.419 & 0.351 & 0.212 & 0.551 & 0.289 \\
VoteFlow & SeFlow & 0.538 & 0.355 & 0.605 & 0.780 & 0.410 & 0.347 & 0.197 & 0.548 & 0.298 \\
DeFlow$++$ & SeFlow$++$ & 0.509 & \underline{0.327} & 0.583 & 0.716 & 0.409 & 0.323 & 0.201 & 0.521 & 0.247 \\
DeltaFlow & TeFlow & \underline{0.433}  & 0.347 & \underline{0.492} & \underline{0.628} & \textbf{0.266} & \underline{0.285} & \underline{0.164} & \textbf{0.486} & \underline{0.206} \\
\rowcolor{lightblue}
DeltaFlow & \method{} (ours) &  \textbf{0.388} & \textbf{0.291} & \textbf{0.319} & \textbf{0.578} & \underline{0.365} & \textbf{0.275} & \textbf{0.124} & \underline{0.507} & \textbf{0.194} \\
\bottomrule
\end{tabular}
}
\end{table}
Besides Argoverse2, we conduct further in-domain tests on nuScenes and Waymo datasets. 
The dynamic normalized EPE on these two datasets is shown in \tab{nusc_waymo_val}. 
Note that the results of TeFlow~\cite{zhang2026teflow} are reproduced using the released pretrained models, whereas the results of other methods are taken from the TeFlow paper, as their checkpoints are not publicly available.
Overall, DeltaFlow+\method{} achieves state-of-the-art performance on both datasets in terms of mean dynamic normalized EPE. 
At the category level, DeltaFlow+\method{} consistently outperforms DeltaFlow+TeFlow on Car and Other Vehicle. 
However, \method{} occasionally underperforms on Pedestrian and Wheeled VRU. 
A possible explanation is that these objects often exhibit non-rigid or articulated motion, which violates the rigid-motion assumption.
This experiment further supports our claim that incorporating motion rigidity into loss design enhances the quality of flow estimation.

\subsection{Ablation Study on Voting Components}
\label{sec:ablation}
\begin{table}[htbp]
\centering
\caption{Ablation of $k$ in voting modules on the AV2 \textit{val} split~\cite{Argoverse2}. (\textbf{Best}, \underline{Second Best})}
\label{tab:ablationtopk}
\setlength{\tabcolsep}{5pt}
\renewcommand{\arraystretch}{0.95}
\resizebox{\columnwidth}{!}
{
\begin{tabular}{c c|c c cc c | c c c c}
\toprule
\multirow{2}{*}{Stage 1} & \multirow{2}{*}{Stage 2}   & \multicolumn{5}{c|}{Dynamic Normalized EPE $\downarrow$} &\multicolumn{4}{c}{Three-way EPE (cm) $\downarrow$} \\
\cmidrule(lr){3-7} \cmidrule(lr){8-11}
& &  Mean & Car & O.V. & Ped. & VRU & Mean & FD & FS & BS \\
\midrule

- & 1 & 0.499 & 0.373 & 0.667 & 0.540 & 0.418 & 7.19 & 18.5 & \textbf{1.81} & \textbf{1.29} \\
3 & 1 & 0.317 & 0.176 & 0.303 & 0.474 & 0.316 & 4.44 & 9.84 & \underline{2.03} & \underline{1.47} \\
5 & 1 & \underline{0.277} & \underline{0.137} & \underline{0.260} & \underline{0.418} & 0.291 & 3.73 & 7.75 & 1.98 & 1.46 \\
10 & 1 & \textbf{0.262 }& \textbf{0.121} & \textbf{0.246} & \textbf{0.405} & \textbf{0.277} & \textbf{3.47} & \textbf{6.84} & 2.05 & 1.53\\
20 & 1 & 0.297 & 0.127 & 0.276 & 0.500 & \underline{0.285} & \underline{3.67} & \underline{7.41} & 2.04 & 1.57\\
\bottomrule
\end{tabular}
}
\end{table}

We validate our design choices using the DeFlow backbone~\cite{zhang2024deflow}.
\tab{ablationtopk} presents the ablation study of the \textit{runoff vote} process, 
including the necessity of two-stage voting and the choice of top-$k$. 
%
The variant that only uses a single stage produces the highest mean error, which is more than 10 p.p. higher than the other two-stage variants, showcasing the demand for two-stage voting.
Empirically, we find that $k=10$ achieves the best performance.
%

\subsection{\protect\rev{Geometric Runoff Voting on VoteFlow Features}}
\label{app:voteflow-probe}
\rev{
We perform a two-stage voting analysis on VoteFlow~\cite{lin2025voteflow}. 
VoteFlow and RVLoss vote over different evidence. 
RVLoss votes over geometric translations, 
whereas VoteFlow’s in-model voting module aggregates cosine similarities between source and target pillar features.
For each source pillar, we inspect VoteFlow’s voting space and calculate the hit rate between the Top-1/Top-10 candidate flows and the ground-truth flows using a threshold $\tau=0.1~\mathrm{m}$, 
which accounts for quantization errors.
\emph{VF Top-1} evaluates the highest-scoring bin in the feature space, whereas
\emph{VF Top-10} measures whether at least one of the ten highest-scoring bins is a hit.}

\rev{Following the two-stage voting principle of \method{}, 
we use the
Top-10 bins as the first-stage translation candidates. 
Their bin centers are decoded into metric translations and used to warp the source points toward the target. 
For each candidate, we perform GRF (Guided Rigid Flow Estimation) under the guidance of the candidate translation. 
The resulting point-level translations then cast unit votes in a geometric voting space, 
where the highest-scoring bin is selected as \emph{RV Top-1}.
As a reference, we also report \emph{VF decoded}, VoteFlow's point-wise flow predictions, 
evaluated on the same points using the same hit criterion.}  

As shown in Tab.~\ref{tab:voteflow-probe}, VF Top-1
achieves a hit rate of $24.2\%$, while VF Top-10 achieves a hit rate of $59.9\%$. 
VoteFlow struggles to identify the optimal candidate, as the performance gap between VF Top-10 and VF Top-1 remains substantial. 
When using the VF Top-10 translation candidates for geometric runoff voting, RV Top-1 consistently improves over VF Top-1 across all categories. This demonstrates that the runoff mechanism can effectively select a better candidate from the given set.

VF Top-10 is an oracle coverage metric rather than a prediction.
The $6.4$ p.p. gain applies to the raw VF Top-1 decision;
it does not establish an improvement or superiority over its full learned voting representation.
This experiment provides mechanistic evidence for geometric runoff
at the intermediate voting stage.
Determining whether it improves final scene-flow predictions requires
integration into VoteFlow, retraining, and controlled evaluation.
\begin{table}[htbp]
\centering
\caption{\protect\rev{\textbf{Voting on frozen VoteFlow feature candidates.}
}}
\label{tab:voteflow-probe}
\small
\rev{\begin{tabular*}{\linewidth}{@{\extracolsep{\fill}}lrrrrr@{}}
\toprule
Category & VF Top-1 (\%) & VF Top-10 (\%) & RV Top-1 (\%) & Gain (p.p.) & VF decoded (\%) \\
\midrule
Car            & 21.5 & 55.2 & 28.0 & $+6.5$  & 65.7 \\
O.V. & 25.2 & 63.2 & 28.1 & $+3.0$  & 48.4 \\
Ped.     & 40.5 & 75.2 & 68.8 & $+28.4$ & 84.1 \\
VRU    & 40.1 & 81.8 & 62.4 & $+22.3$ & 57.7 \\
\midrule
All foreground & 24.2 & 59.9 & 30.6 & $+6.4$  & 59.5 \\
\bottomrule
\end{tabular*}}
\end{table}

\begin{table}[!b]
\centering
\caption{\method{} as offline Flow Refinement. We use \method{} to refine the predicted flows and thus obtain rigid flows and free-form flows. We compare these flows on the AV2 \emph{val} split~\cite{Argoverse2} using mean dynamic normalized EPE. (\textbf{Best}/\underline{Second Best} in each row and category)
}
\label{tab:flow_refinement}
\setlength{\tabcolsep}{5pt}
\renewcommand{\arraystretch}{0.95}
\resizebox{\columnwidth}{!}
{
\begin{tabular}{l c|c c cc c | c c cc c | c c cc c}
\toprule
\multirow{2}{*}{Method} & \multirow{2}{*}{Loss}   & \multicolumn{5}{c|}{Predicted flow$\downarrow$} &\multicolumn{5}{c|}{(Refined) rigid flow$\downarrow$}  &\multicolumn{5}{c}{(Refined) free-form flow$\downarrow$} \\
\cmidrule(lr){3-7} \cmidrule(lr){8-12} \cmidrule(lr){13-17}
& &  Mean & Car & O.V. & Ped. & VRU &  Mean & Car & O.V. & Ped. & VRU &  Mean & Car & O.V. & Ped. & VRU\\
\midrule
DeFlow & SeFlow &  0.342 & 0.205 & 0.340 & 0.510 & 0.315 & \textbf{0.237} & \textbf{0.140} & \textbf{0.273} & \textbf{0.348} & \textbf{0.189} & \underline{0.293} & \underline{0.194} & \underline{0.294} & \underline{0.411} & \underline{0.272}  \\
DeFlow$++$ & SeFlow$++$ & 0.300 & 0.180 & 0.340 & 0.395 & 0.288 & \textbf{0.236} & \textbf{0.131} & \textbf{0.260} & \textbf{0.363} & \textbf{0.189} & \underline{0.289} & \underline{0.187} & \underline{0.282} & \underline{0.416} & \underline{0.270}  \\
DeltaFlow & TeFlow & \underline{0.223} & 0.140 & 0.268 & \textbf{0.304} & \underline{0.179} & \textbf{0.220} & \textbf{0.119} & \textbf{0.243} & \underline{0.354} & \textbf{0.162} & 0.278 & 0.178 & 0.267 & 0.414 & 0.255 \\
\bottomrule
\end{tabular}
}
\end{table}
\subsection{\method{} for offline flow refinement }

Optionally, the runoff vote process in \method{} can also be used as an offline refinement module that adjusts the predicted flow from a pretrained model, producing both refined rigid flow and refined free-form flow. We compare the predicted and refined flows in \tab{flow_refinement}.
For Car and Other Vehicle, the refined rigid flow consistently improves over the predicted flow, demonstrating the benefit of \textit{runoff vote} in enhancing motion rigidity. 
However, it may degrade performance for Pedestrian, highlighting its limitations in handling non-rigid motion.
The refined free-form flow exhibits varying behavior depending on the quality of the initial predictions. For DeFlow and DeFlow++ models, it generally improves over the predicted flow. In contrast, when applied to DeltaFlow, the refined free-form flow no longer yields gains, suggesting that it is approaching an upper performance bound (mean dynamic normalized EPE around 0.28).

\subsection{\protect\rev{Stage-wise Analysis of Pseudo-label Construction}}
\label{app:stage-diagnostics}

We analyze \method{}'s pseudo-label construction
using DeFlow trained with \method{} on AV2 \emph{val} split.
We compare the model prediction, first-stage rigid flow candidates,
second-stage rigid flow, and final free-form flow against the
GT flow. Mean dynamic normalized EPE (Dynamic Mean) is reported.   

\paragraph{Stage definitions.}

\emph{Model prediction} is the direct output of the DeFlow backbone.
\emph{Top-$k$ oracle flow} evaluates the candidate set produced
by the first voting stage
(Sec.~\ref{sec:topk_voting}), with $k=10$.
Specifically, we select a single candidate flow per cluster that minimizes the mean 3D endpoint distance to the GT flow. 
The selected candidate is shared by all points within the cluster.
\emph{Top-1 rigid flow} is the cluster-shared translation selected
by the second voting stage
(Sec.~\ref{sec:top1_voting}).
\emph{Free-form flow} is the subsequent point-wise flow obtained
through nearest-neighbor retrieval
(Sec.~\ref{sec:pseudo_label}).


\begin{table}[t]
\centering
\caption{\protect\rev{
\textbf{Stage-wise pseudo-label accuracy on AV2 \emph{val} split.}
$\Delta_{\mathrm{sel}}$ measures the selection gap between
second-stage voting and the GT-assisted reference
(Top-1 $-$ Top-$k$ oracle).
$\Delta_{\mathrm{ret}}$ measures the error change after final
point-wise nearest-neighbor retrieval (Free-form $-$ Top-1).
Positive differences indicate degradation.
}}
\label{tab:app-stage-diagnostics}
\small
\setlength{\tabcolsep}{3pt}
\renewcommand{\arraystretch}{1.15}
\rev{\begin{tabular*}{\textwidth}{@{\extracolsep{\fill}}lrrrrrr@{}}
\toprule
Category & Model prediction & Top-$k$ oracle & Top-1 rigid & Free-form flow & $\Delta_{\mathrm{sel}}$ & $\Delta_{\mathrm{ret}}$ \\
\midrule
Car & 0.121 & 0.067 & 0.109 & 0.171 & +0.042 & +0.062 \\
O.V. & 0.246 & 0.123 & 0.221 & 0.246 & +0.098 & +0.025 \\
Ped. & 0.407 & 0.287 & 0.345 & 0.409 & +0.058 & +0.064 \\
VRU & 0.275 & 0.132 & 0.176 & 0.260 & +0.044 & +0.084 \\

\midrule
\textbf{Dynamic Mean} & 0.262 & 0.152 & 0.213 & 0.272 & +0.061 & +0.059 \\
\bottomrule
\end{tabular*}}
\end{table}

\paragraph{\protect\rev{Overall behavior.}}

\rev{Tab.~\ref{tab:app-stage-diagnostics} shows Top-$k$ oracle reduces mean dynamic normalized EPE
from $0.262$ to $0.152$, indicating that the first-stage
candidate set contains valid flow hypotheses for each cluster.
Second-stage selection yields a mean dynamic normalized EPE of $0.213$,
and free-form retrieval further increases it to $0.272$.
The dominant source of diagnostic error differs by category.}

\paragraph{\protect\rev{Pedestrian.}}
\rev{Pedestrian retains the highest Top-$k$ oracle error among the evaluated categories (0.287), even when ground truth is used to select the best cluster-shared candidate. This indicates that its performance gap is already present in the first-stage voting. This behavior is consistent with articulated or heterogeneous intra-cluster pedestrian motion. The subsequent top-1 selection and free-form retrieval introduce additional increases of 0.058 and 0.064, respectively. We therefore attribute the gap on the Pedestrian category primarily to limited candidate/shared-motion representation.
}

\paragraph{\protect\rev{Wheeled VRU.}}
\rev{In contrast, degradation on the Wheeled VRU category comes mostly from the free-form retrieval, increasing the error by 0.084 -- the largest such increase among the evaluated categories. This identifies point-wise nearest-neighbor retrieval as the principal bottleneck for Wheeled VRU. Sparse and thin geometry, partial visibility, and rider–vehicle motion differences may make cross-frame point correspondences ambiguous, although isolating their individual effects requires further analysis.}

\paragraph{\protect\rev{Discussion.}}
\rev{These results show that the underperforming categories do not share a single failure mode: performance on Pedestrian is primarily influenced by the Top-$k$ rigid flow candidates, whereas performance on Wheeled VRU is primarily influenced by the final free-form retrieval.
}

\subsection{\protect\rev{Runtime Analysis of Runoff Voting}}
\label{app:runtime}
\rev{
We analyze a detailed runtime breakdown of the two-stage runoff-voting procedure for $k\in\{3,5,10,20\}$
(Tab.~\ref{tab:app-runtime}). 
The \texorpdfstring{$1^{\text{st}}$}{1st} GRF represents the \texorpdfstring{$1^{\text{st}}$}{1st} stage voting for top-k rigid flows,
which uses the predicted flow to find nearest neighbors, then votes for $k$
candidate translations. 
The \texorpdfstring{$2^{\text{nd}}$}{2nd} GRF is the \texorpdfstring{$2^{\text{nd}}$}{2nd} stage voting for top-1 rigid flow,
which repeats the nearest-neighbor search
for each candidate and votes for one translation per cluster. 
A final nearest-neighbor search guided by this translation yields the free-form flow.}
\rev{We profile 20 AV2 validation pairs on an NVIDIA A100 80GB PCIe GPU, using two
warm-up and seven measured repetitions for each $k$. For each stage, we
sum forward and inverse CUDA-event times per repetition.
Timing covers pseudo-label construction, excluding the backbone, loss
reduction, backpropagation, and data loading.}

\begin{table}[t]
\centering
\caption{\protect\rev{\textbf{Runtime breakdown.}
Times are milliseconds per bidirectional frame pair. \texorpdfstring{$2^{\text{nd}}$}{2nd} GRF share is the \texorpdfstring{$2^{\text{nd}}$}{2nd} GRF time divided by the Sum column.
Dynamic Mean is the mean dynamic normalized EPE from the separate training ablation in Tab.~\ref{tab:ablationtopk}.}}
\label{tab:app-runtime}
\small
\setlength{\tabcolsep}{3pt}
\rev{\begin{tabular*}{\linewidth}{@{\extracolsep{\fill}}rrrrrrr@{}}
\toprule
$k$ & \texorpdfstring{$1^{\text{st}}$}{1st} GRF & \texorpdfstring{$2^{\text{nd}}$}{2nd} GRF & Free-form KNN & Sum & \texorpdfstring{$2^{\text{nd}}$}{2nd} GRF share & Dynamic Mean $\downarrow$ \\
\midrule
3 & 22.16 & 60.14 & 22.00 & 104.30 & 57.7\% & 0.317 \\
5 & 22.35 & 96.79 & 21.98 & 141.13 & 68.6\% & 0.277 \\
10 & 22.61 & 188.25 & 21.92 & 232.77 & 80.9\% & \textbf{0.262} \\
20 & 22.34 & 352.71 & 21.84 & 396.90 & 88.9\% & 0.297 \\
\bottomrule
\end{tabular*}}
\end{table}

\rev{At the default $k=10$, the runtimes sum to 232.77\,ms per
bidirectional frame pair. The \texorpdfstring{$2^{\text{nd}}$}{2nd} GRF accounts for 80.9\% of this
sum and is therefore the primary computational bottleneck.  
Its runtime grows approximately linearly with $k$, increasing from
60.14 to 352.71\,ms as $k$ increases from 3 to 20, consistent with
performing a nearest-neighbor search for each candidate branch.
The free-form
retrieval remains nearly constant and takes 21.84--22.00\, ms across these settings, which introduces only a marginal additional cost.
Importantly, this cost is incurred only during training, when the runoff procedure constructs pseudo-labels for the self-supervised objective. Neither runoff voting nor its additional nearest-neighbor searches are executed during standard inference. Consequently, RVLoss introduces no additional inference-time operations or latency and leaves the inference procedure of the underlying scene-flow backbone unchanged.}

\subsection{Broader Impact}
We introduce a self-supervised loss for scene flow estimation that is compatible with existing feedforward backbones. By removing the need for manual annotations, it has the potential to substantially reduce data collection and labeling costs. At the same time, it improves scalability by enabling training on large-scale unlabeled LiDAR datasets commonly available in autonomous driving.
More accurate dynamic scene understanding can enhance the safety and reliability of point cloud-based perception systems, particularly in autonomous driving and robotics applications. However, increased scalability may come at the cost of higher computational demands, potentially raising the carbon footprint of model training. In addition, there is a risk that scene flow models could be misused for unauthorized tracking or surveillance.
To mitigate these risks, it is essential to adopt responsible data governance, enforce access controls, and adhere to established ethical principles, ensuring the technology is deployed for societal benefit.



\end{document}